\documentclass[sigconf]{acmart}
\AtBeginDocument{%
  }

\copyrightyear{2026}
\acmYear{2026}
\setcopyright{cc}
\setcctype{by}
\acmConference[KDD '26]{Proceedings of the 32nd ACM SIGKDD Conference on Knowledge Discovery and Data Mining V.2}{August 09--13, 2026}{Jeju Island, Republic of Korea}
\acmBooktitle{Proceedings of the 32nd ACM SIGKDD Conference on Knowledge Discovery and Data Mining V.2 (KDD '26), August 09--13, 2026, Jeju Island, Republic of Korea}
\acmDOI{10.1145/3770855.3817680}
\acmISBN{979-8-4007-2259-2/2026/08}
\usepackage{booktabs}
\usepackage{array}
\usepackage{graphicx}
\usepackage[normalem]{ulem}
\usepackage{multirow}
\usepackage{enumitem}
\usepackage{tabularray}
\usepackage[dvipsnames]{xcolor}
\usepackage{pifont}
\usepackage{subcaption}
\usepackage{adjustbox}
\usepackage{caption}
\usepackage{rotating}
\usepackage{diagbox}
\usepackage{tabu}
\usepackage{xspace}
\usepackage[utf8]{inputenc}
\usepackage{geometry}
\usepackage[most]{tcolorbox}

\newtcolorbox{promptbox}[1][]{
  enhanced,
  breakable,
  title=\textbf{System Prompt for MultiDocR Question Generation},
  colback=gray!5!white,
  colframe=gray!75!black,
  coltitle=white,
  fonttitle=\bfseries\large,
  sharp corners,
  boxrule=0.8pt,
  #1
}

\newtcolorbox{ocrbox}[1][]{
  enhanced,
  breakable,
  title=\textbf{OCR Extraction Sample},
  colback=gray!5!white,
  colframe=gray!60!black,
  coltitle=white,
  fonttitle=\bfseries\large,
  fontupper=\small\sffamily,
  sharp corners,
  boxrule=0.8pt,
  left=2mm, right=2mm, top=2mm, bottom=2mm,
  #1
}

\newcommand{\cmark}{\textcolor{Green}{\ding{51}}}
\newcommand{\xmark}{\textcolor{Red}{\ding{55}}}

\begin{document}

\title{DocRetriever: A Plug-and-Play Framework for Multimodal Document Retrieval with Comprehensive Benchmark}

\author{Ruofan Hu}
\authornote{Equal contribution.}
\affiliation{%
  \institution{Zhejiang University}
  \city{Hangzhou}
  \state{Zhejiang}
  \country{China}
}
\email{ruofanhu@zju.edu.cn}
\orcid{0009-0000-4618-1140}

\author{Menghui Zhu}
\authornotemark[1]
\affiliation{%
  \institution{Huawei Technologies Co., Ltd}
  \city{Shanghai}
  \state{Shanghai}
  \country{China}
}
\email{zhumenghui1@huawei.com}
\orcid{0000-0002-8567-2185}

\author{Jieming Zhu}
\authornote{Corresponding Authors.}
\affiliation{%
  \institution{Huawei Technologies Co., Ltd}
  \city{Shenzhen}
  \state{Guangdong}
  \country{China}
}
\email{jamie.zhu@huawei.com}
\orcid{0000-0002-5666-8320}

\author{Bo Chen}
\affiliation{%
  \institution{Huawei Technologies Co., Ltd}
  \city{Shanghai}
  \state{Shanghai}
  \country{China}
}
\email{chenbo.31@qq.com}
\orcid{0000-0003-3750-2533}

\author{Shengyang Xu}
\affiliation{%
  \institution{Zhejiang University}
  \city{Hangzhou}
  \state{Zhejiang}
  \country{China}
}
\email{3230104220@zju.edu.cn}
\orcid{0009-0002-5705-7932}

\author{Minjie Hong}
\affiliation{%
  \institution{Zhejiang University}
  \city{Hangzhou}
  \state{Zhejiang}
  \country{China}
}
\email{hongminjie@zju.edu.cn}
\orcid{0009-0000-0368-2527}

\author{Xiaoda Yang}
\affiliation{%
  \institution{Zhejiang University}
  \city{Hangzhou}
  \state{Zhejiang}
  \country{China}
}
\email{1992426088@qq.com}
\orcid{0009-0002-7297-4536}

\author{Sashuai Zhou}
\affiliation{%
  \institution{Zhejiang University}
  \city{Hangzhou}
  \state{Zhejiang}
  \country{China}
}
\email{22421039@zju.edu.cn}
\orcid{0009-0004-9245-4639}

\author{Li Tang}
\affiliation{%
  \institution{Zhejiang University}
  \city{Hangzhou}
  \state{Zhejiang}
  \country{China}
}
\email{tanglzju@zju.edu.cn}
\orcid{0009-0001-0461-8452}

\author{Tao Jin}
\affiliation{%
  \institution{Zhejiang University}
  \city{Hangzhou}
  \state{Zhejiang}
  \country{China}
}
\email{jint_zju@zju.edu.cn}
\orcid{0000-0003-3564-1628}

\author{Zhou Zhao}
\authornotemark[2]
\affiliation{%
  \institution{Zhejiang University}
  \city{Hangzhou}
  \state{Zhejiang}
  \country{China}
}
\email{zhaozhou@zju.edu.cn}
\orcid{0000-0001-6121-0384}

\renewcommand{\shortauthors}{Hu et al.}

\begin{abstract}

Multimodal documents contain diverse elements, such as tables, figures, and layouts, which can complicate retrieval tasks. While current approaches typically combine dense visual embedding models with supervised rerankers to achieve high-precision retrieval, they face inherent limitations. First, the coarse-grained nature of dense embeddings tends to obfuscate explicit semantics, failing to leverage structurally salient information. Second, supervised reranking models suffer from generalization bottlenecks, as their performance heavily relies on domain-specific training data. Furthermore, existing benchmarks often lack diverse assessment dimensions and comprehensive relevance annotations, limiting reliable evaluation. To address these challenges, we propose DocRetriever, a plug-and-play framework. It enhances visual retrieval via a layout-aware sparse embedding technique, enabling effective hybrid encoding without the overhead of optical character recognition (OCR). We also introduce a generalizable reranker that leverages reasoning-augmented demonstrations and optimized sampling to improve accuracy in few-shot settings. Finally, we construct a new benchmark, MultiDocR, to enable more rigorous evaluation. Experiments across diverse benchmarks validate DocRetriever's superiority over state-of-the-art methods.

\end{abstract}



\maketitle

\section{Introduction}

\begin{table*}[t]
\renewcommand\arraystretch{0.9} 
\centering
\small
\resizebox{0.8\linewidth}{!}{
\begin{tabular}{l@{\hskip 4pt}|c@{\hskip 3pt}c@{\hskip 3pt}c|c@{\hskip 4.5pt}c@{\hskip 4.5pt}c@{\hskip 4.5pt}c|c@{\hskip 2pt}c@{\hskip 2pt}c}
\toprule
\multirow{2}{*}{\textbf{Benchmarks}}  
  & \multicolumn{3}{c|}{\textbf{Document}} 
  & \multicolumn{4}{c|}{\textbf{Question}} 
  & \multicolumn{3}{c}{\textbf{Eval. Metric}} \\
& Domain & \#Pages & Source & \#Num & Expert & Tagged & Rew. & Ans. & Page. & Verified \\
\midrule
MP-DocVQA \cite{tito2023hierarchical} 
  & Industrial & 8.3 & \cmark & 46k & \xmark & \xmark & \xmark & \cmark & \cmark & \xmark \\
PDF-MVQA \cite{ding2024mvqa}  
  & Biomedical & 9.6 & \cmark & 260k & \xmark & \xmark & \xmark & \cmark & \cmark & \xmark \\
DocBench \cite{zou2024docbench} 
  & Multiple & 66.0 & \xmark & 1,102 & \cmark & \cmark & \xmark & \xmark & \xmark & \cmark \\
MMLongBench-Doc \cite{ma2024mmlongbench}  
  & Multiple & 47.5 & \cmark & 1,082 & \xmark & \xmark & \xmark & \cmark & \cmark & \xmark \\
Wiki-SS \cite{ma2024unifying} 
  & Wikipedia & 1.0 & \xmark & 3,610 & \xmark & \xmark & \xmark & \xmark & \cmark & \xmark \\
ViDoRe \cite{faysse2024colpali} 
  & Multiple & 1.0 & \xmark & 3,810 & \xmark & \xmark & \xmark & \xmark & \cmark & \xmark \\
MMDocIR \cite{dong2025mmdocir}  
  & Multiple & 65.1 & \cmark & 1,658 & \cmark & \xmark & \xmark & \cmark & \cmark & \xmark \\
MMDocRAG \cite{dong2025benchmarking} 
  & Multiple & 67.0 & \cmark & 4,055 & \cmark & \cmark & \xmark & \cmark & \cmark & \xmark \\
M3DocVQA \cite{cho2025m3docvqa} 
  & Multiple & 67.0 & \cmark & 4,055 & \cmark & \cmark & \xmark & \cmark & \cmark & \xmark \\
\midrule
\normalsize MultiDocR 
  & Multiple & 12.2 & \cmark & 2,441 & \cmark & \cmark & \cmark & \cmark & \xmark & \xmark \\
\bottomrule
\end{tabular}
}
\caption{Comparison between MultiDocR and existing DocVQA/DocIR benchmarks: \textit{Source}: Document provenance traceable; \textit{Expert}: Questions expert-verified; \textit{Tagged}: Query type taxonomy; \textit{Rew.}: Query reformulation for lexical robustness; \textit{Ans.}: Text answers provided; \textit{Page.}: Retrieval targets provided; \textit{Verified}: Retrieval relevance validated.}

\label{tab:dataset_comparison}
\end{table*}

Document retrieval is a cornerstone of modern information retrieval systems, enabling users to locate relevant information from large-scale multimodal corpora encompassing PDFs, web pages, and posters. In contrast to methods \cite{zhang2022multi, chen2023walking} which primarily deal with textual passages, multimodal document retrieval requires a more holistic understanding of visual elements including images, tables and layout designs. This challenge has led to research progress along two key directions: (1) embedding models that map queries and documents into a unified latent space for efficient similarity computation, and (2) reranking mechanisms capable of performing fine-grained, multimodal relevance assessment.

Recent embedding methods for document retrieval follow two main paradigms. The first employs vision-language models (VLMs) to transform multimodal inputs into structured text, which are encoded via textual dense embedding \cite{karpukhin2020dense, chen2024bge, wang2024multilingual, izacard2021unsupervised, li2023towards}, bag-of-words approaches \cite{robertson2009probabilistic, ramos2003using, formal2021splade}, or recent hybrid approaches like PromptReps \cite{zhuang2024promptreps} that combine both capabilities. However, this process relies on costly extraction pipelines (e.g., ~29.4 s/page for Qwen2.5VL-32B in MMDocIR \cite{dong2025mmdocir, bai2025qwen2}), where the information loss also compromises embedding fidelity \cite{dong2025benchmarking,hu2025vela}. In contrast, the second paradigm treats document pages as snapshots, using VLMs to aggregate embeddings from hidden states \cite{yu2024visrag, faysse2024colpali, dse}. While efficient, these methods fail to preserve explicit word-level alignments between queries and structurally salient document regions, which are critical for robust generalization \cite{lesk1969word, aalbersberg1994document, bastiaanse2016role}. Although attempts like MLSR \cite{nguyen2024multimodal} aim to recover lexical semantics via generated summaries, they reintroduce substantial computational overhead.

The reranking stage faces similar limitations. Mirroring the embedding paradigm, prevalent methods employ textual rerankers on document passages \cite{li2023making, chen2024bge, zhang2025qwen3, zhang2024mgte}, thereby risking the omission of visual semantics. To mitigate this, recent work has proposed fine-tuned VLMs that jointly encode queries and document screenshots, scoring relevance by the probability of a "Yes" response \cite{MonoQwen, gunther2025jina, xu2025mm}. However, constrained by the scarcity of high-quality open-source datasets \cite{mathew2021docvqa,li2025mrg,cheng2025voxdialogue}, these models often suffer from limited generalization to out-of-distribution documents or unseen query types, especially those requiring complex layout-aware reasoning beyond simple pattern matching.

Given these challenges, we propose \textbf{DocRetriever}, a plug-and-play framework that enhances existing VLM-based retrieval systems through three key innovations:

First, we propose a layout-aware sparse embedding technique to construct the hybrid-encoding architecture. While VLM-based embedding systems typically encode final hidden states into dense embeddings via pretrained projection layers, the language modeling (LM) head inherently generates a vocabulary-scale distribution with logits reflecting semantic importance. Leveraging this distribution, we derive chunk-adaptive sparse embeddings through optimized frequency-based reweighting and top-256 logit selection. Integrating these sparse embeddings into our hybrid framework yields a consistent 3\% improvement in NDCG@10 over the dense embedding baseline.

Second, we propose an In-Context Learning (ICL) framework to enhance reranking generalization. While ICL effectively adapts VLMs to new scenarios, conventional approaches are hindered by their reliance on costly manually curated demonstrations. To address this, we introduce a Reinforced ICL strategy, enabling the autonomous synthesis of reasoning-augmented demonstrations via rigorous cross-verification. During inference, we leverage dual query-document similarity to retrieve the most pertinent examples, demonstrating superior results across diverse benchmarks.

Lastly, we introduce MultiDocR, a benchmark designed for rigorous evaluation. Existing datasets often assume a strict one-to-one correspondence between queries and answer pages. However, document information is frequently redundant across pages and modalities. This widespread redundancy leads to incomplete annotations, where semantically relevant pages are often mislabeled as negatives. Moreover, current benchmarks lack sufficient domain and query diversity, limiting multi-dimensional evaluation. To address this, MultiDocR spans 10 document domains and 7 query categories (Tab. 2). It also incorporates query paraphrases to test lexical robustness and provides verified text extractions. Finally, we replace binary labels with five-level relevance scores, enabling a more nuanced and comprehensive assessment.

In summary, the contributions of this work are as follows:

\begin{itemize}[leftmargin=*]
    \item We propose a plug-and-play hybrid encoding mechanism for VLM-based retrievers. By extracting layout-aware sparse embeddings directly from the LM head's logit distribution, our method achieves superior dense-sparse fusion and retrieval accuracy without incurring additional OCR costs.
    
    \item We introduce a Reinforced ICL framework for document reranking. Through autonomous demonstration synthesis and a dual-similarity sampling strategy, our reranker effectively mitigates data scarcity issues, achieving robust generalization without the need for fine-tuning.
    
    \item We construct MultiDocR, a comprehensive benchmark featuring multi-dimensional query taxonomies, lexical paraphrases, and fine-grained 5-level relevance annotations. This resource addresses the limitations of existing binary-labeled datasets, enabling a more rigorous and systematic evaluation of document retrieval systems.
\end{itemize}

\section{Related Work}

\begin{figure*}[htbp]
    \centering
    \includegraphics[page=1, width=0.85\linewidth, clip, trim=195 150 195 125]{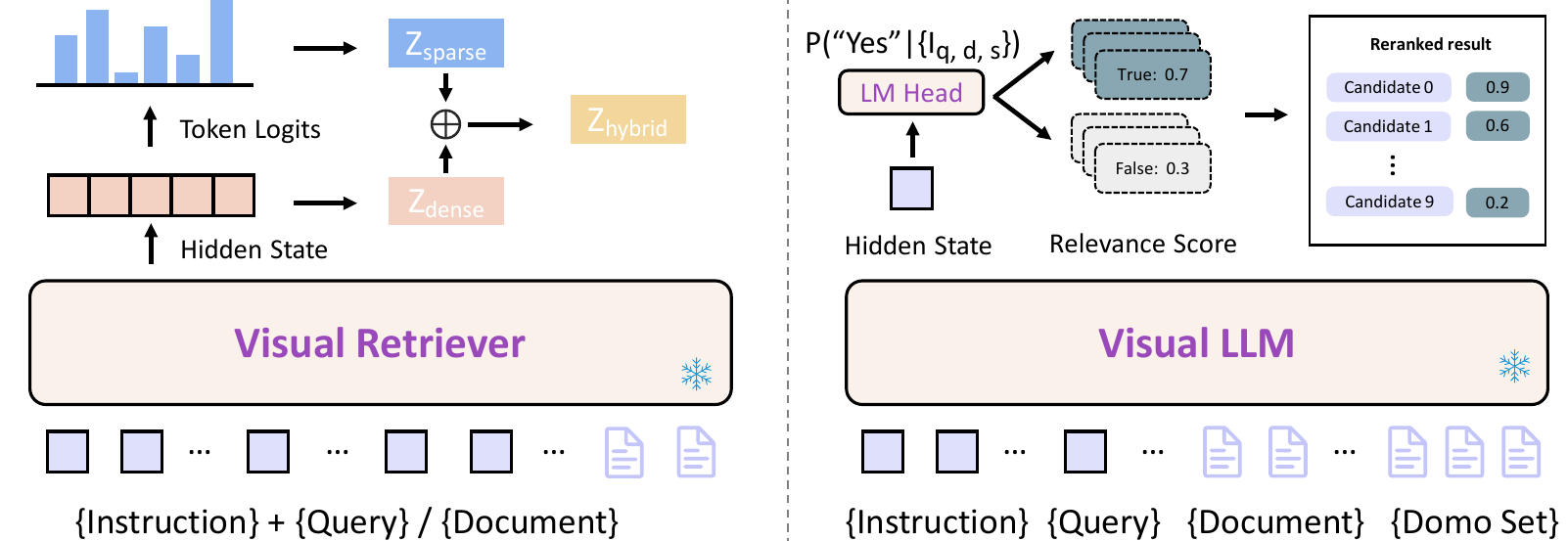}
    \caption{Model architecture of our Hybrid Encoding (left) and Reranker with ICL (right).}
    \label{fig:main}
\end{figure*}

\subsection{Multimodal Document Retrieval Benchmark}
\label{sec:2.1}
Multimodal document retrieval benchmarks often stem from Document VQA datasets focused on perceptual understanding. Key examples range from single-page datasets like DocVQA \cite{mathew2021docvqa}, InfoVQA \cite{mathew2022infographicvqa}, and TAT-DQA \cite{zhu2022towards}, to multi-page extensions including DUDE \cite{van2023document}, MP-DocVQA \cite{tito2023hierarchical}, SlideVQA \cite{tanaka2023slidevqa} and MR$^2$-Bench \cite{zhou2025mr}. Recent works, such as MMLongBench-Doc \cite{ma2024mmlongbench}, DocBench \cite{zou2024docbench}, M-LongDoc \cite{chia2024m} and GlobalQA \cite{luo2025towards}, further target entire documents, thereby introducing substantial noise from irrelevant content.

In document retrieval research, some studies \cite{faysse2024colpali, yu2024visrag} adapt these DocVQA subsets by treating questions as queries and target pages as relevant items. Recent work \cite{dong2025mmdocir, dong2025benchmarking,mace2025vidore} further refines these by filtering unsuitable queries and documents lacking contextual clarity. In contrast, other approaches \cite{dse} bypass existing VQA datasets altogether and construct corpora from raw sources like Wikipedia. However, these benchmarks typically adhere to a strict one-to-one mapping between queries and ground-truth pages, overlooking the inherent redundancy of information that may span or be re-expressed across multiple pages. This limitation compromises evaluation reliability. Moreover, retrieval performance often fluctuates based on query phrasing \cite{lei2024meta,tao2024llms}, question type \cite{li2025mrg}, and document domain \cite{dong2025benchmarking}, making it difficult to disentangle model capability from benchmark artifacts. These gaps collectively highlight the need for a more comprehensive and systematic evaluation framework. To our knowledge, no existing benchmark systematically combines multi-dimensional evaluation with rigorous query-content alignment.

\subsection{Embedding-Reranking Ecosystem for Document Retrieval}
Document retrieval systems adopt a two-stage pipeline: embedding-based retrieval followed by neural reranking. The first stage evolved from sparse lexical methods like TF-IDF \cite{ramos2003using,zhuohao2021keyword}, BM25 \cite{robertson2009probabilistic}, and SPLADE \cite{formal2021splade} to neural textual dual-encoders \cite{reimers2019sentence, karpukhin2020dense, wang2022text,fang2025gta} for dense semantic search. Building upon these developments, hybrid retrieval methods like PromptReps \cite{mandikal2024sparse,zhuang2024promptreps,nguyen2024multimodal,yang2025multimodal,hong2025generative,hu2025vela} combine lexical and semantic embeddings, consistently outperforming standalone paradigms.

Nevertheless, these text-centric methods incur substantial latency due to extraction pipelines and often fail to fully leverage native visual layout features. To address this, recent studies shift toward direct visual encoding via VLMs, generating dense embeddings through mean pooling or [CLS] token aggregation. Specifically, Document Screenshot Embedding (DSE) \cite{dse}, VisRAG \cite{yu2024visrag} and VLM2Vec \cite{jiang2024vlm2vec} map entire pages into a single embedding space. ColPali and ColQwen \cite{faysse2024colpali} further advance this by segmenting page images into chunks to preserve fine-grained details. However, these paradigms struggle to integrate sparse lexical features, thereby failing to achieve the hybrid encoding efficacy of text-based systems.

To further refine the precision, modern systems employ neural rerankers on retrieved candidates. Early encoder-based rerankers, such as BERT \cite{devlin2019bert}, utilize bidirectional attention for exhaustive query-document interaction modeling. Generative approaches like MonoT5 \cite{lin2024mm,liu2025llm4ranking} shifted this paradigm, leveraging language model token logits for relevance scoring. Recent models like Qwen3-Reranker~\cite{zhang2025qwen3} and Jina-Reranker-m0~\cite{gunther2025jina} further validate generative reranking's efficacy. However, scarcity of high-quality open-domain data and the alignment issues noted in Sec. \ref{sec:2.1} constrain the generalization and robustness of these models. 

\section{Method}

\subsection{Overview}
We provide an overview of DocRetriever in Fig. \ref{fig:main}. Given a query $Q$ and a corpus $\mathcal{D} = \{D_i\}_{i=1}^n$, DocRetriever aims to retrieve a candidate subset $\mathcal{D}^+_Q = \{ D^+_{Q,j} \}_{j=1}^m \subset \mathcal{D}$, where $m \ll n$ denotes the number of retrieved documents. Specifically, in the offline phase, documents are encoded by VLMs to derive both dense and sparse embeddings from the visual input. Next, during online inference, the system computes the hybrid embedding for $Q$, enabling modality-aligned similarity matching against the pre-encoded document embeddings. These dense and sparse similarity scores are independently normalized and combined via a weighted sum to yield a unified ranking, from which the top-$m$ documents are retained.

In the reranking stage, DocRetriever leverages fine-grained cross-modal interactions to process the query $Q$ and candidate set $\mathcal{D}^+_Q$, thereby yielding a re-ordered ranking $\hat{\mathcal{D}}^+_Q$ with higher-fidelity relevance assessment. Specifically, we adopt a \textbf{pointwise} reranking strategy where the VLM independently computes a scalar relevance score for each query-document pair $(Q, D^+_{Q,i})$. To enhance reranking accuracy and generalization, DocRetriever utilizes ICL, where the input is prepended with reasoning-augmented demonstrations to guide more precise scoring.

\subsection{Embedding for Hybrid Retrieval}
\label{sec:3.2}
DocRetriever augments standard VLM-based dense retrieval models with sparse embeddings to enhance retrieval precision. While previous frameworks effectively combine dense and sparse embeddings for text-only retrieval, their application to visually-rich documents necessitates costly and error-prone pipelines to extract lexical features \cite{dong2025mmdocir,dong2025benchmarking}. To address this, DocRetriever extracts sparse embeddings directly from the VLM's LM head alongside dense ones, enabling a unified and OCR-free process. Importantly, to ensure plug-and-play compatibility with diverse document retrieval backbones, two tailored variants are designed. 

The first approach, exemplified by VisRAG \cite{yu2024visrag, dse,jiang2024vlm2vec}, employs prompt engineering to compress the entire page into the final layer hidden states ($h$). Specifically, we modify their prompt from Represent this document:' to Represent this document in one word:'. This concise instruction yields superior sparsity, guiding these models to aggregate document semantics into the final token while preserving dense representation quality \cite{tao2024llms, lei2024meta, thirukovalluru2024geneol}. This hidden state, when projected through the LM head ($p$), yields a discriminative vocabulary-scale logit distribution $v$ that captures semantic importance via sparse token weights.
 
The second approach, exemplified by ColPali and ColQwen \cite{faysse2024colpali}, processes documents via chunk-wise encoding to facilitate late-interaction mechanisms. Unlike the first approach, this segmented strategy localizes semantic information within individual chunks. We independently encode each chunk, projecting the final hidden states ($h$) through the LM head to derive a discriminative distribution over the vocabulary. To consolidate these $M$ chunk-level representations, max-pooling is applied across the vocabulary dimension, producing a final distribution $v$ that retains the most semantically salient lexical features from the entire document.

However, the raw logit distribution $v \in \mathbb{R}^{|\mathcal{V}|}$ derived from the LM Head is inherently dense, assigning non-zero weights to the entire vocabulary. To transform $v$ into retrieval-efficient sparse embeddings, DocRetriever employs a three-step sparsification pipeline:

\begin{enumerate}[leftmargin=*]
    \item \textbf{Word Lemmatization}: Leveraging the NLTK toolkit \cite{loper2002nltk}, we perform part-of-speech tagging and lemmatization \cite{plisson2004rule} to consolidate lexical variations. Specifically, logits of inflected forms (e.g., \textit{running}, \textit{runner}, \textit{ran}) are aggregated to their corresponding base lemma (e.g., \textit{run}) by retaining the maximum value.
    
    \item \textbf{Token Filtering}: To reduce noise, standard stopwords and invalid characters are filtered out from the candidate set.

    \item \textbf{Logit Processing}: 
    Following SPLADE \cite{formal2021splade}, ReLU activation and log-saturation are applied to the logits to mitigate the dominance of high-frequency tokens. The resulting weights are subsequently truncated to the top 256 dimensions and then scaled by a factor of 100 for integer quantization.
\end{enumerate}

Finally, after obtaining the effective sparse embedding, DocRetriever computes the similarity between the query and documents in both the sparse and original dense embedding spaces. These similarity scores are then normalized and combined via a weighted sum to obtain the hybrid retrieval score:

\begin{equation}
\begin{split}
    S_{\text{hybrid}}(D, Q) 
    &= \lambda \cdot \text{Sim}_{norm}(z_{\text{dense}}(D), z_{\text{dense}}(Q)) \\
    &\quad + (1 - \lambda) \cdot \text{Sim}_{norm}(z_{\text{sparse}}(D), z_{\text{sparse}}(Q))  ~,
\end{split}
\label{eq:eq2} 
\end{equation}

where $z_{\text{dense}}$ and $z_{\text{sparse}}$ denote the dense and sparse embedding functions and $\text{Sim}_{\text{norm}}$ is the min-max normalized cosine similarity measuring the angle between embeddings. Based on \cite{mandikal2024sparse}, we fix $\lambda = 0.8$ for optimal performance.

\subsection{Reinforced ICL for Reranker}
\label{sec:Reinforced ICL for Reranker}

\begin{figure*}[t]
    \begin{adjustbox}{valign=t,minipage=0.48\linewidth}
        \centering
        \renewcommand\tabcolsep{2.2pt}
        \renewcommand\arraystretch{0.82}
        \resizebox{\linewidth}{!}{%
        \begin{tabular}{lc}
        \toprule
        \textbf{Statistic} & \textbf{Number} \\
        \midrule
        \textbf{Documents}   & 313  \\
        \quad - Domain Types      & 10 \\
        \quad - Avg./Med./Max. pages per doc & 65 / 26 / 843 \\
        \quad - Avg./Med./Max. words per doc & 23k / 7k / 260k \\
        \quad - Avg./Med./Max. images per doc & 18 / 9 / 53 \\
        \quad - Avg./Med./Max. tables per doc & 16 / 11 / 42 \\
        \midrule
        \textbf{Total Questions}   & 2,581  \\
        \quad - Derived questions & 1,535 (59.4\%) \\ 
        \quad - Newly-annotated questions & 1,046 (40.6\%) \\
        \quad - Cross-page questions &  1,132 (43.8\%) \\
        \quad - Cross-modal questions & 1,372  (53.2\%) \\
    
        \quad - Question types & 7 \\
        \multicolumn{2}{l}{%
          \hspace*{1em}
          \begin{tabular}{@{\quad}l@{\quad}r@{\qquad}l@{\quad}r}
            Analytical: & 432 (16.8\%) & Comparative: & 489 (18.9\%) \\
            Descriptive: & 534 (20.7\%) & Explanatory: & 440 (17.0\%) \\
            Procedural: & 426 (16.5\%) & Others: & 260 (10.1\%)
          \end{tabular}%
        } \\
        
        \midrule
        \textbf{Relationship} & \\
        \quad - Raw Page Related to Query & 2.35 : 1 \\
        \quad - Annotated Page Related to Query & 3.94 : 1 \\
        \quad - Evidence distribution \\
        \multicolumn{2}{l}{%
          \hspace*{1em}
            \begin{tabular}{@{\quad}lr@{\quad}lr}
                Text: & 60.4\%  \quad\quad\quad\quad\quad\quad\quad & Image: & 18.8\% \\
                Table: & 16.7\% \quad\quad\quad\quad\quad\quad\quad & Other: & 4.1\% \\
            \end{tabular}%
        } \\
    
        \bottomrule
        \end{tabular}%
        }
        \captionof{table}{Overall dataset statistics for MultiDocR.}
        \label{tab:dataset_stats}
    \end{adjustbox}
    \hfill
    \begin{adjustbox}{valign=t,minipage=0.475\linewidth}
        \centering
        \includegraphics[width=\linewidth, clip, trim=35mm 36.1mm 35mm 36mm, page=1]{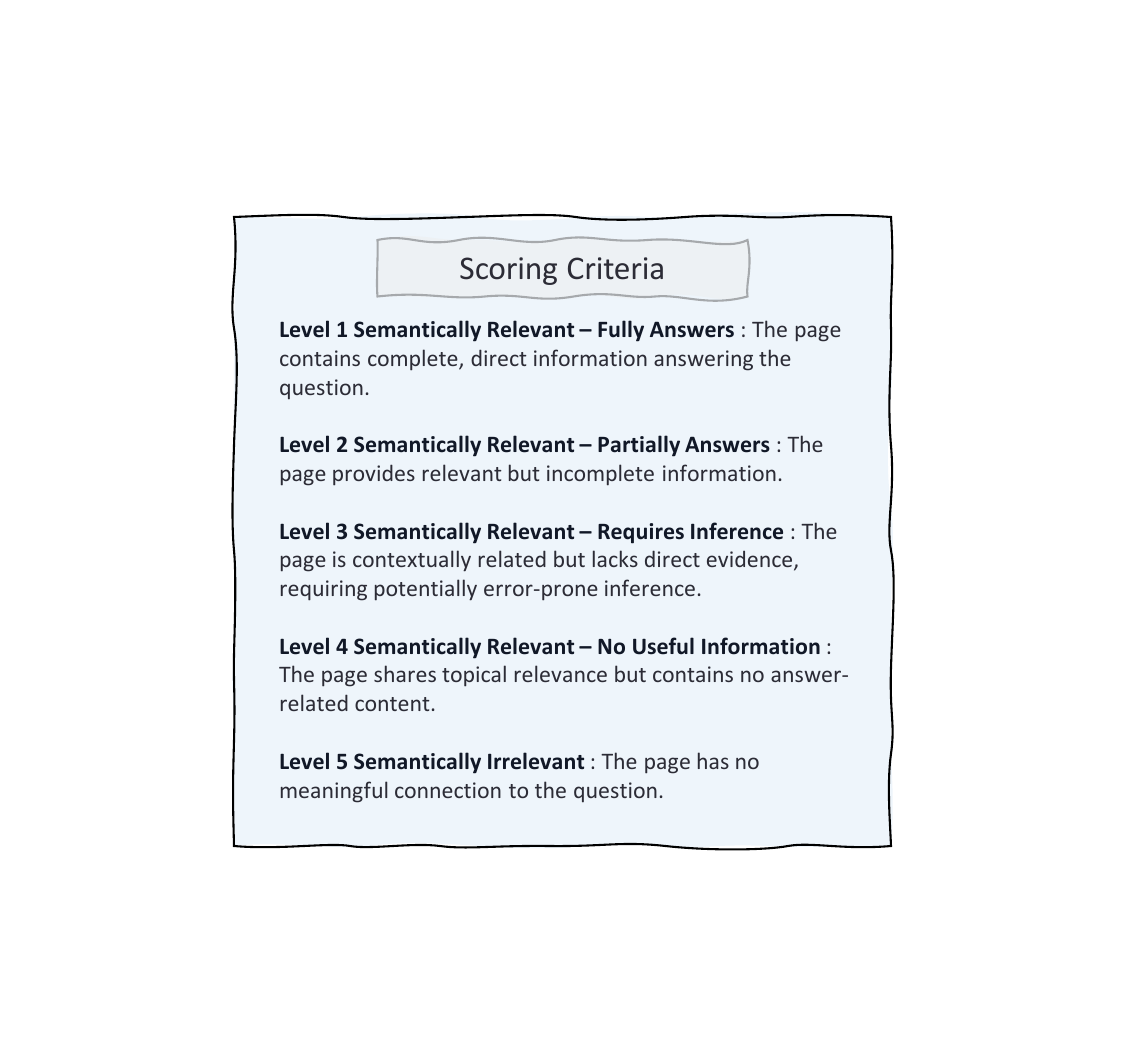}
        \caption{Score Distribution.}
        \label{fig:score}
    \end{adjustbox}
\end{figure*}

For the reranking stage, DocRetriever employs a reinforced ICL framework to enhance performance. Document reranking is typically formulated as a binary classification task, where VLMs estimate relevance based on the logit of a target token \cite{nogueira2020document}. While fine-tuning on retrieval datasets can enhance performance, it often leads the model to overfit to dataset-specific artifacts, resulting in poor robustness when transferring to unfamiliar retrieval scenarios. Furthermore, the scarcity of fine-grained, multi-level annotations in existing datasets limits effective optimization, leading to poor generalization across diverse domains. These challenges motivate our ICL framework, which serves as a data-efficient alternative that adapts VLMs to new scenarios by dynamically incorporating reasoning-augmented demonstrations. 
 
Generally, effective ICL demonstrations comprise three key components: (1) task instruction, (2) step-by-step reasoning chains, and (3) target relevance label. Current paradigms frequently leverage large-scale frontier VLMs to synthesize these components, allowing lower-parameter models to benefit from enhanced reasoning capabilities without fine-tuning. However, despite their scale, these teacher models often exhibit a pronounced positivity bias during complex relevance assessments \cite{wang2024large}. This bias can introduce seemingly plausible yet logically flawed reasoning chains into the demonstration set, ultimately degrading reranking performance.

To mitigate this issue, DocRetriever employs a contrastive verification strategy. Instead of generating examples individually, we instruct the VLM to produce reasoning chains for a pair of ground-truth positive ($d^+$) and high-scoring hard negative ($d^-$) samples for the same query. A generated demonstration is included only if the VLM yields correct relevance predictions for both samples in the pair. Additionally, to ensure logical fidelity, we employ a multi-model consensus protocol with an ensemble of frontier models \cite{comanici2025gemini, achiam2023gpt, yang2025qwen3}. For each instance, one model proposes a reasoning chain while peer models act as reviewers to strictly filter out inconsistent rationales. By leveraging such cross-verification, we effectively mitigate model biases and hallucinations, achieving a 98.3\% precision rate in a manual audit of 3,000 samples. Detailed hyperparameters are provided in App.~\ref{sec:Reinforced ICL Details and Qualitative Analysis} for reproducibility.

Moreover, standard ICL relies exclusively on textual query similarity, often failing to capture the visual cues essential for generalization in document reranking. To address this, DocRetriever employs a dual-alignment strategy that integrates query semantics with document visual similarity. Based on our hybrid embeddings, we retrieve the most similar demonstrations based on a joint metric of query and document similarity, which is validated in Sec. \ref{analysis} to yield optimal performance compared to uni-modal approaches. These examples will condition the VLM's prediction:

\begin{equation}
    \text{score}(q, d) = \frac{e^{P(\text{Yes} | I_{q,d,s})}}{e^{P(\text{Yes} | I_{q,d,s})} + e^{P(\text{No} | I_{q,d,s})}} ~,
\label{eq:eq3} 
\end{equation}

where $I_{q,d,s}$ denotes the instruction, query, document pages and context from the demo set, and $P(\cdot\mid\cdot)$ represents the VLM's conditional probability distribution.

\subsection{MultiDocR}
\label{sec:MultiDocR}
Finally, we introduce MultiDocR, a comprehensive retrieval benchmark extending MMDocIR \cite{dong2025mmdocir}. The base benchmark contains 313 long-form multimodal documents, 1,658 expert-annotated questions, and OCR-extracted text for each page. The diverse visual elements within document pages make it particularly suitable for evaluating retrieval systems.

However, MMDocIR still exhibits several limitations. First, it assumes a one-to-one correspondence between queries and pages. In practice, relevant information often spans multiple pages, necessitating a multi-page relevance paradigm to accurately evaluate the retrieval of scattered evidence. Second, the benchmark focuses on document domains but lacks query type classification, which is a critical component of comprehensive evaluation. Third, MMDocIR’s queries often replicate exact phrases from the documents, whereas real-world searches frequently require matching synonyms or paraphrased expressions. 

To address these issues, Gemini-2.5 pro \cite{comanici2025gemini} is employed to systematicallyrefine and expand the benchmark into MultiDocR via the following procedure:

\textbf{Diversity Enhancement.}
Queries are classified into seven fundamental question types: analytical, comparative, descriptive, explanatory, inferential, procedural, and regulatory. Noting the significant imbalances in both question categories and document domains across MMDocIR, 1,463 new queries with paired pages and answers are generated to create a more balanced set. To minimize superficial pattern matching, each query is also rephrased to achieve low lexical overlap (Jaccard similarity = 0.15) while preserving semantic equivalence with the supporting evidence.

\textbf{Data Curation.}
The 3,121 questions undergo a rigorous review process, with three quality filters applied: (1) removal of query-item pairs solvable through text-only retrieval without visual information, (2) elimination of questions requiring complex external knowledge, and (3) filtration of overly vague questions. This process results in a final set of 2,581 high-quality questions.

\textbf{Relevance Annotation.}
The relevance assessment follows a hierarchical three-stage process. First, given the large corpus size, ColQwen's \cite{faysse2024colpali} hybrid retrieval capability is leveraged to identify 50 candidate pages per query. Second, this pool is subsequently narrowed to the top 30 most relevant pages via Jina-Reranker-m0 \cite{gunther2025jina}. Finally, to transcend the limitations of traditional binary relevance, Gemini2.5-pro is employed with a five-level scale (Fig. \ref{fig:score}) to assign the relevance score for these pages. This annotation design enables a more precise evaluation through weighted nDCG@10, effectively rewarding retrievers that prioritize highly informative pages over those with only partial relevance.

Key statistics for MultiDocR are reported in Tab.~\ref{tab:dataset_stats}. Ethical considerations including privacy and bias mitigation are discussed in App.~\ref{sec:Ethical Considerations}.

\section{Experiment}
\begin{table*}[t] 
\small
\setlength{\tabcolsep}{2.5pt}
\renewcommand{\arraystretch}{0.8}
    \centering
    \resizebox{0.88\linewidth}{!}{%
  \begin{tabu}{lll|cccccccccc|c}

    \toprule

    \multicolumn{3}{c}{\multirow{2}{*}{\diagbox{Method}{Domain}}} 
     & Resear. & Admin & Tutori.\& & Acade. & \xspace Broch- & Finance & Guide- & Govern- & \multirow{2}{*}{\xspace Laws \xspace} & \multirow{2}{*}{\xspace News\xspace}  & \multirow{2}{*}{Average} \\
     & & & Report & \&Indu. & Worksh. & Paper & ure & Report & book & ment & & & \xspace \xspace \\
    \midrule
   
    \parbox[t]{2.5mm}{\multirow{25}{*}{\rotatebox[origin=c]{90}{nDCG@$k=10$}}} &
    \parbox[t]{2.0mm}{\multirow{3}{*}{\rotatebox[origin=c]{90}{Lex.}}}
     & TF-IDF & 52.6 & 47.4 & 32.3 & 42.2 & 34.6 & 23.2 & 44.8 & 44.3 & 47.9 & 44.4 & 40.4 \\
    && BM25 & 46.6 & 41.7 & 15.7 & 44.1 & 26.4 & 21.3 & 45.8 & 35.1 & 50.5 & 41.4 & 36.9 \\
    && SPLADE & 52.3 & 48.5 & 22.1 & 49.8 & 36.4 & 31.2 & 51.5 & 45.1 & 58.7 & 46.4 & 44.2 \\
    \tabucline[.2pt on 1pt off 1.5pt]{2-14} \\[-5pt]

    & \parbox[t]{1.5mm}{\multirow{5}{*}{\rotatebox[origin=c]{90}{Sparse}}}
    & ColQwen  & 43.3 & 63.2 & 59.7 & 51.5 & 58.6 & 34.5 & 54.3 & 57.2 & 64.4 & 36.4 & 52.3 \\
    && ColPali & 22.7 & 28.1 & 20.1 & 21.7 & 22.1 & 19.6 & 14.6 & 22.3 & 23.2 & 24.1 & 21.8 \\
    && DSE$_{\mathrm{docmatix}}$  & 43.1 & 56.0 & 41.4 & 41.1 & 44.8 & 40.1 & 38.4 & 38.0 & 51.4 & 41.0 & 43.5 \\
    && VisRAG & 50.0 & 37.1 & 39.3 & 39.8 & 48.9 & 33.8 & 35.0 & 49.4 & 48.3 & 35.2 & 41.6 \\
    && VLM2Vec & 38.0 & 37.1 & 30.6 & 25.7 & 33.8 & 27.2 & 32.0 & 44.5 & 37.4 & 17.7 & 32.4 \\
    \tabucline[.2pt on 1pt off 1.5pt]{2-14} \\[-5pt]

    & \parbox[t]{2.0mm}{\multirow{5}{*}{\rotatebox[origin=c]{90}{Text}}}
     & BGE & 64.9 & 60.2 & 48.3 & 41.3 & 52.4 & 37.0 & 52.8 & 55.9 & 58.1 & 24.8 & 49.6 \\
    && E5 & 62.2 & 57.1 & 48.3 & 43.7 & 51.8 & 43.6 & 55.5 & 51.3 & 61.6 & 26.4 & 50.2 \\
    && Contriever & 62.9 & 56.3 & 45.8 & 38.0 & 56.6 & 32.4 & 52.3 & 52.0 & 58.2 & 25.1 & 48.0 \\
    && GTE & 63.4 & 61.7 & 49.1 & 41.5 & 50.6 & 37.1 & 51.9 & 56.9 & 58.0 & 23.2 & 49.3 \\
    && Qwen3$_{\mathrm{Embed.}}$ & 70.4 & 63.9 & 54.7 & 54.4 & 57.7 & 51.3 & 60.9 & 66.0 & 64.7 & 37.4 & 58.1 \\
    \tabucline[.2pt on 1pt off 1.5pt]{2-14} \\[-5pt]

    & \parbox[t]{1.5mm}{\multirow{5}{*}{\rotatebox[origin=c]{90}{Dense}}}
    & ColQwen & 82.8 & 77.7 & 74.2 & 72.8 & 71.8 & 70.2 & 74.8 & 75.8 & 76.6 & 52.2 & 72.9 \\
    && ColPali  & 80.0 & 76.5 & 72.6 & 65.7 & 69.6 & 67.5 & 71.6 & 74.3 & 73.7 & 42.7 & 69.5 \\
    && DSE$_{\mathrm{docmatix}}$  & 70.4 & 77.1 & 67.3 & 63.0 & 63.3 & 60.8 & 65.6 & 69.8 & 73.7 & 51.9 & 66.5 \\
    && VisRAG & 75.5 & 67.6 & 67.9 & 58.6 & 61.6 & 63.1 & 59.0 & 68.1 & 64.7 & 42.9 & 62.9 \\
    && VLM2Vec & 74.0 & 69.9 & 66.1 & 64.4 & 62.4 & 60.8 & 65.3 & 67.9 & 68.2 & 44.0 & 64.3 \\
    \tabucline[.2pt on 1pt off 1.5pt]{2-14} \\[-5pt]

    & \parbox[t]{1.5mm}{\multirow{2}{*}{\rotatebox[origin=c]{90}{Hy.}}}
    & PromptReps & 51.1 & 46.2 & 47.4 & 45.6 & 47.8 & 51.8 & 46.6 & 42.1 & 51.5 & 47.8 & 52.4 \\
    && MLSR & 68.8 & 63.5 & 63.8 & 53.6 & 56.7 & 59.8 & 54.6 & 64.1 & 58.3 & 38.8 & 58.2 \\
    \tabucline[.2pt on 1pt off 1.5pt]{2-14} \\[-5pt]
    
    & \parbox[t]{1.5mm}{\multirow{5}{*}{\rotatebox[origin=c]{90}{Ours}}}
    & ColQwen & \textbf{84.1} & \textbf{80.2} & \underline{76.5} & \textbf{74.8} & \underline{75.9} & \textbf{74.3} & \textbf{77.2} & \textbf{78.6} & \textbf{79.2} & \textbf{56.7} & \textbf{75.7} \\
    && ColPali & 83.0 & 77.6 & 73.8 & 66.7 & 70.2 & 69.1 & 72.3 & 75.5 & 75.7 & 45.1 & 70.8 \\
    && DSE$_{\mathrm{docmatix}}$ & 72.6 & 78.3 & 68.8 & 66.5 & 65.6 & 63.2 & 68.6 & 72.2 & 75.8 & \underline{56.1} & 68.8 \\
    && VisRAG & 76.9 & 71.6 & 71.9 & 61.7 & 64.8 & 67.8 & 62.6 & 72.1 & 66.3 & 46.8 & 66.2 \\
    && VLM2Vec & 77.1 & 74.0 & 69.6 & 67.6 & 65.9 & 64.2 & 70.2 & 73.1 & 72.7 & 50.1 & 67.8 \\

    \bottomrule
  \end{tabu}
  }

\caption{Retrieval performance (NDCG@10) on the MultiDocR benchmark across different document domains. The best and second-best scores are highlighted in bold and \underline{underlined}, respectively. Methods are categorized as: \textit{Lex.}: Lexical sparse baselines; \textit{Text}: Textual dense baselines; \textit{Sparse}: Our constructed sparse embeddings; \textit{Dense}: VLM-native dense embeddings; \textit{Hy.}: Hybrid embedding baselines; and \textit{Ours}: Our proposed hybrid embedding.}
\label{tab:main_page_recall}
\end{table*}
\subsection{Experiment Setup}
We conduct comprehensive evaluations on MultiDocR and other established benchmarks to assess the proposed retrieval pipeline.

We evaluate (1) text-based sparse retrievers: BM25, TF-IDF, SPL-ADE (110M); (2) text-based dense retrievers: BGE (335M), E5 (335M), Contriever (109M), GTE (335M), Qwen3-Embedding (8B); (3) VLM-based dense retrievers: VisRAG (3B), DSE (4B), ColQwen (2B), ColPali (3B), VLM2Vec (4B). For VLMs, DocRetriever augments their native dense embeddings with constructed sparse embeddings to form a hybrid representation as Eq. \eqref{eq:eq2}. We also benchmark PromptReps (8B), a text-based hybrid model, and MLSR (8B), which integrates textual document summaries with visual dense embeddings.

For the reranking stage, we evaluate diverse baselines, spanning text-based models such as Bge-reranker (567M), Qwen3-reranker (8B), GTE-reranker (305M), and Jina-multilingual-reranker (278M), to visual rerankers like Jina-reranker-m0 (2B), MonoQwen2-VL (2B) and MM-R5 (8B). DocRetriever employs Qwen2.5VL-7B-Instruct as the backbone. To ensure a strict zero-shot setting, ICL demonstrations are drawn exclusively from the MMDocIR training set, removing any document ID overlaps with external benchmarks.

In experiments, weighted nDCG@10 is adopted \cite{wang2013theoretical} as the primary evaluation metric, as detailed in App. \ref{sec:Evaluation Metric Definition}. For text-based models, the OCR-extracted content is provided as input, whereas VLM-based methods operate directly on document screenshots.

\subsection{Main Result}
\subsubsection{Experiment on Hybrid Embedding}

\label{Hybrid Embedding}
We evaluate DocRetriever on MultiDocR, presenting domain-level results.

\textbf{Sparse Embedding. } As shown in Tab. \ref{tab:main_page_recall}, our sparse embedding (Sparse) consistently rivals or surpasses traditional lexical approaches (Lex.) across all domains. This advantage arises from fundamental differences in term weighting. Specifically, while traditional bag-of-words models depend solely on term frequency and exact lexical matching, DocRetriever integrates layout-aware semantics, a critical factor for robust document understanding. For example, terms in structurally salient regions (e.g., section headers or figure captions) often carry greater semantic importance despite their lower frequency. DocRetriever leverages visual layout cues to adaptively upweight these salient terms, thereby significantly enhancing retrieval precision.

Moreover, we compare our sparse embedding method (Sparse) with text-based dense retrieval models (Text). As shown in the same table, our sparse embedding achieves competitive performance, and in several cases even exceeds that of dense models. This parity stems from the VLM's ability to implicitly learn term relationships: semantically synonymous terms corresponding to salient words will be assigned higher weights, even when such terms are lexically absent in documents (e.g., "2D" reinforcing "animation"). Such behavior is crucial for real-world retrieval, where queries often paraphrase rather than repeat exact terms in the document. Furthermore, this behavior mirrors the synonym robustness of dense retrieval models, where lexically divergent but semantically related terms cluster in latent space. DocRetriever’s sparse logits replicate this effect at the token level, activating similar weights for related tokens while retaining high interpretability.
 
\textbf{Hybrid Embedding. } For the hybrid embedding (Ours), we follow the implementation described in Sec. \ref{sec:3.2}. As shown in Tab. \ref{tab:main_page_recall}, this approach consistently improves nDCG scores by approximately 3\% compared to visual dense encoding alone (Dense).Specifically, the VLM-based hybrid approach significantly surpasses PromptReps, underscoring the critical role of visual modalities in complex multimodal document retrieval (see App.~\ref{sec:Methodological Comparisons Clarifications} for a detailed comparison). Conversely, although MLSR includes a visual encoder, its reliance on pre-extracted text summaries negates the advantages of layout-aware modeling, diminishing retrieval performance due to the loss of fine-grained structural information. These findings are further supported by end-to-end RAG experiments and generalization on external benchmarks in App.~\ref{sec:End-to-End RAG Performance} and App.~\ref{sec:Generalization on External Benchmarks}. Notably, we observe that certain base models (e.g., ColPali) yield weaker standalone sparse representations, and we further analyze this phenomenon in Sec. \ref{analysis}.

\begin{table*}[t] 
\small
\setlength{\tabcolsep}{2.5pt}
\renewcommand{\arraystretch}{0.8}
    \centering
    \resizebox{0.95\linewidth}{!}{%
  \begin{tabu}{lll|cccccccccc|cc}

    \toprule

    \multicolumn{3}{c}{\multirow{2}{*}{\diagbox{Method}{Domain}}} 
     & Resear. & Admin & Tutori.\& & Acade. & \xspace Broch- & Finance & Guide- & Govern- & \multirow{2}{*}{\xspace Laws \xspace} & \multirow{2}{*}{\xspace News\xspace}  & \multirow{2}{*}{Average} & \multirow{2}{*}{Latency} \\
     & & & Report & \&Indu. & Worksh. & Paper & ure & Report & book & ment & & & \xspace \xspace & \\
    \midrule
   
    \parbox[t]{2.5mm}{\multirow{11}{*}{\rotatebox[origin=c]{90}{nDCG@$k=10$}}} &

    \parbox[t]{2.0mm}{\multirow{1}{*}{\rotatebox[origin=c]{90}{}}}
     & ColQ$_{\mathrm{hy}}$ & 84.1 & 80.2 & 76.5 & 74.8 & 75.9 & 74.3 & 77.2 & 78.6 & 79.2 & 56.7 & 75.6 & - \\
    \tabucline[.2pt on 1pt off 1.5pt]{2-15} \\[-5pt]

    & \parbox[t]{2.0mm}{\multirow{4}{*}{\rotatebox[origin=c]{90}{Text}}}
     & BGE$_{\mathrm{reranker}}$ & 72.9 & 69.5 & 57.8 & 63.0 & 64.2 & 66.4 & 73.3 & 70.3 & 74.8 & 33.0 & 64.5 & 6.5 \\
    && GTE$_{\mathrm{reranker}}$ & 77.7 & 74.5 & 58.8 & 71.1 & 69.6 & 73.3 & 77.5 & 74.2 & 76.2 & 34.7 & 68.7 & 3.9 \\
    && Jina$_{\mathrm{reranker-v2}}$ & 75.3 & 68.3 & 58.8 & 63.2 & 65.9 & 66.8 & 72.9 & 69.6 & 71.8 & 33.7 & 64.6 & 4.2 \\
    && Qwen3$_{\mathrm{reranker}}$ & 81.9 & 77.3 & 64.7 & 79.4 & 68.7 & 83.4 & 80.5 & 79.5 & 83.5 & 64.3 & 76.3 & 43.6 \\
    \tabucline[.2pt on 1pt off 1.5pt]{2-15} \\[-5pt]

    & \parbox[t]{1.5mm}{\multirow{2}{*}{\rotatebox[origin=c]{90}{Vis.}}}
    & Jina$_{\mathrm{raranker-m0}}$  & \textbf{92.7} & \underline{89.4} & \underline{87.5} & \textbf{89.0} & \textbf{87.0} & \underline{85.4} & \textbf{86.1} & \underline{87.6} & \textbf{91.6} & \underline{70.4} & \underline{86.7} & 14.6 \\
    && MM-R5  & 84.2 & 85.1 & 86.8 & 78.2 & 83.0 & 76.8 & \underline{85.3} & 81.9 & 85.6 & 66.4 & 81.2 & 49.5 \\
    && MonoQwen2  & 87.3 & 83.0 & 83.3 & 83.2 & 84.3 & 78.2 & 83.4 & 83.1 & 89.3 & 69.1 & 82.6 & 13.5 \\
    \tabucline[.2pt on 1pt off 1.5pt]{2-15} \\[-5pt]

    & \parbox[t]{2.0mm}{\multirow{3}{*}{\rotatebox[origin=c]{90}{ICL}}}
     & Random & 85.1 & 81.3 & 83.6 & 74.8 & 82.9 & 84.3 & 76.4 & 86.0 & 84.2 & 65.1 & 80.5 & 33.4 \\
    && Difficult & 84.4 & 82.2 & 82.1 & 71.2 & 83.9 & 83.1 & 77.2 & 85.4 & 82.5 & 66.2 & 79.7 & 33.4 \\
    && Similar & \underline{92.4} & \textbf{90.9} & \textbf{90.7} & \underline{86.6} & \underline{85.7} & \textbf{90.2} & 83.2 & \textbf{91.8} & \underline{91.0} & \textbf{74.8} & \textbf{87.8} & 33.4 \\
    
    \bottomrule
  \end{tabu}
  }
\caption{Main results of NDCG@10 for reranking. Latency denotes the average inference time in seconds, excluding OCR transcription costs of textual baselines. The ICL example retrieval adds negligible overhead ($\sim$ 2 ms) during online inference.}
\label{tab:rerank}
\end{table*}

\textbf{Domain Analysis. } The analysis also reveals consistent underperformance across all dense models on the News domain. Through systematic investigation, we identify two primary contributing factors: (1) the prevalence of domain-specific terminology and (2) the dispersion across pages. These findings highlight critical limitations in current retrieval architectures. First, the model's encoding process often fails to capture low-frequency terms due to their insufficient presence in the training corpus, which hinders robust representation learning. Second, although some VLM-based approaches process pages in chunks, they lack mechanisms to effectively model inter-chunk relationships, which limits their ability to handle long-range contextual dependencies. 

However, DocRetriever overcomes these limitations through its sparse embedding approach, which effectively identifies and weights layout-salient terms in documents to produce a highly discriminative similarity matrix. Consequently, the hybrid encoding achieves notable improvements over pure dense embedding in the News domain, with an average enhancement of 5.9\%.

\subsubsection{Experiment on Reranker}
\label{Experiment on Reranker}
We evaluate reranking performance across diverse document domains and query types. For conciseness, we present domain-specific results here.

\textbf{Mechanism of Context Selection.} We investigate three demonstration selection strategies for the ICL framework: (1) Random: denotes selecting four examples from the demonstration pool at random for each iteration; (2) Difficult: identifies and consistently utilizes the four examples with the lowest confidence scores to provide the model with more discriminative information; (3) Similar: dynamically retrieves four examples that exhibit the highest semantic and visual similarity to the target query-document pair.

As shown in Tab. \ref{tab:rerank}, the Similar strategy proves superior, achieving an average nDCG@10 of 87.8 and outperforming the Random baseline by nearly 8 points. This advantage is particularly distinct in the complex \textbf{News} domain. We attribute this performance leap to the topological alignment enabled by the ``Similar'' strategy.  By retrieving examples that are both semantically relevant to the query and visually similar to the document, the model can better internalize the task-specific reasoning logic and effectively apply it to the target pair.

\textbf{Efficiency and Trade-off Analysis.} Tab. \ref{tab:rerank} also contrasts the end-to-end latency of various reranking models when processing the top-30 candidate documents. The results demonstrate that DocRetriever achieves an optimal latency-accuracy trade-off.

Specifically, while most lightweight text-based rerankers demonstrate minimal latency, their limited precision undermines the practical utility in real-world deployment. In contrast, while the large-scale Qwen3-Reranker attains comparable accuracy, it exhibits prohibitive latency (43.6s). This bottleneck arises from the token inflation inherent in text-based representations. The transcription of document images into textual renditions ($\sim$2,500 words) results in sequences of approximately 3,000 tokens, which imposes a non-negligible computational burden on downstream inference. Consequently, text-based rerankers fail to strike a sustainable balance between ranking precision and computational efficiency.

This efficiency-performance mismatch extends to visual rerankers as well. For instance, MM-R5 exhibits prohibitive latency (49.5s) stemming from its reliance on extensive reasoning, rendering it impractical for online inference. Furthermore, while efficient counterparts like MonoQwen2-VL and Jina-Reranker-m0 maintain a competitive latency, they still struggle with limited generalization, particularly in the News domain. In contrast, DocRetriever strikes a more effective balance, yielding superior precision and generalization robustness with a manageable computational cost. First, reasoning-augmented demonstrations enable DocRetriever to outperform baselines by 4.4\% in the News domain, effectively addressing the generalization limits of conventional rerankers. Furthermore, the incremental overhead of $\sim$0.66s per image remains well within the permissible threshold for practical deployment, ensuring its viability for online inference scenarios.

\section{Analysis and Discussion}
\label{analysis}

\paragraph{\textbf{Sparse Embeddings for Different VLMs}}
In Sec. \ref{Hybrid Embedding}, we observe that despite identical training on the same dataset, a significant discrepancy exists between the sparse embeddings produced by the PaliGemma-based ColPali and the Qwen2-VL-based ColQwen. This phenomenon prompts us to investigate whether the base model can influence the lexical distribution of the sparse embedding. To ensure a fair comparison, we evaluate the vanilla pre-trained models using the prompt "in one word:" to guide the compression of information into a single token distribution.

\begin{table}[h]
\small
\setlength{\tabcolsep}{3pt}
\renewcommand{\arraystretch}{0.9}
\centering
\begin{tabular}{l|cccc|c}
\toprule
\multirow{2}{*}{\diagbox{Model}{Domain}} 
 & Resear. & Acade. & \multirow{2}{*}{\xspace News\xspace} & \multirow{2}{*}{Others} 
 & \multirow{2}{*}{Average} \\
 & Report & Paper & & & \\
\midrule
Phi & \underline{9.0} & \underline{4.1} & 1.0 & 3.5 & \underline{4.5} \\
Qwen & \textbf{13.3} & \textbf{7.7} & \textbf{2.3} & \textbf{7.6} & \textbf{7.7} \\
Llama & 4.5 & 0.4 & 0.0 & \underline{2.1} & 2.1 \\
PaliGemma & 8.5 & 2.5 & \underline{1.2} & 3.1 & 3.3 \\

\bottomrule
\end{tabular}
\caption{Performance comparison across different VLMs.}
\label{tab:words distribution}
\end{table}

\begin{figure}[htbp]
    \centering
    \includegraphics[page=2, width=0.99\linewidth, clip, trim=230 116 95 128]{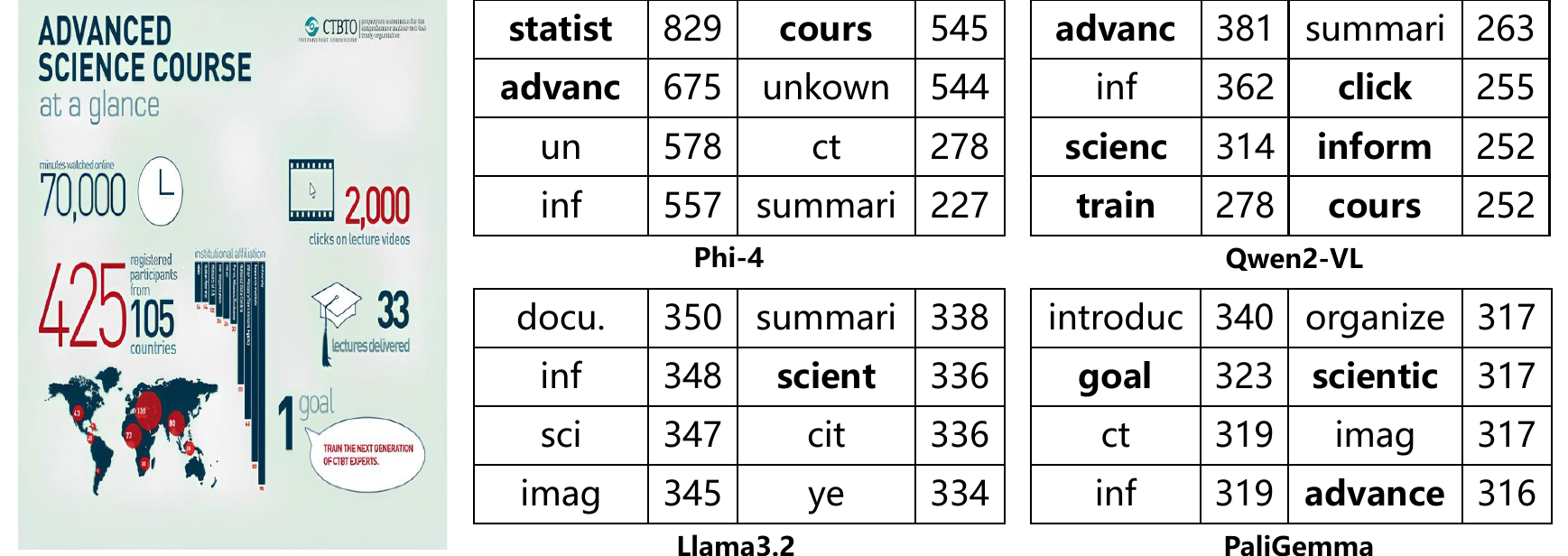}
    \caption{Token distribution from different VLMs.}
    \label{fig:token distribution}
\end{figure}

As shown in Fig. \ref{fig:token distribution} and Tab. \ref{tab:words distribution}, Qwen2-VL (7B) and Phi4 (6B) \cite{abdin2024phi} demonstrate superior performance, effectively extracting meaningful vocabulary while reflecting word importance through highly discriminative weight values. In contrast, PaliGemma (3B) \cite{beyer2024paligemma} identifies fewer valid terms with an overly uniform weight distribution, which undermines sparse embedding efficacy. The Llama3.2 (11B) \cite{touvron2023llama} performs the weakest, with its extracted vocabulary failing to form a meaningful distribution. These results highlight that selecting base models with robust representational capabilities (e.g., Qwen) is critical for effective hybrid retrieval, as this foundation significantly amplifies the performance of VLM-based retrievers.

\paragraph{\textbf{Methods for Semantic Extraction. }}
The key to our efficient sparse embedding extraction lies in achieving proper dimensionality reduction of the vocabulary distribution space. Beyond our proposed method, two alternative paradigms are commonly used to process vocabulary-scale distributions: (1) employing independently trained mapping layers for dimensionality transformation \cite{chen2023stair, li2024unified}, and (2) utilizing the VLM’s input embedding matrix to project the vocabulary distribution space $\mathbb{R}^{|\mathcal{V}|}$ back into a latent semantic space $\mathbb{R}^d$ \cite{hrinchuk2019tensorized,yang2021careful}. Using ColQwen as a backbone, we conduct comparative experiments between these paradigms, specifically implementing a dedicated mapping layer via contrastive learning on the MMDocIR training dataset and a direct parametric projection using the model’s native embedding layer.

\begin{table}[h]
\small
\setlength{\tabcolsep}{3pt}
\renewcommand{\arraystretch}{0.9}
\centering
\begin{tabular}{l|cccc|c}
\toprule
\multirow{2}{*}{\diagbox{Model}{Domain}} 
 & Resear. & Acade. & \multirow{2}{*}{\xspace News\xspace} & \multirow{2}{*}{Others} 
 & \multirow{2}{*}{Average} \\
 & Report & Paper & & & \\
\midrule
Linear Head & \underline{27.3} & \underline{39.8} & \underline{21.3} & \underline{30.6} & \underline{29.8} \\
Embedding Layer & 4.2 & 3.1 & 1.9 & 3.7 & 3.2 \\
Ours & \textbf{33.3} & \textbf{51.5} & \textbf{26.4} & \textbf{53.1} & \textbf{50.3} \\
\bottomrule
\end{tabular}
\caption{Performance comparison across mapping methods.}
\label{tab:mapping methods}
\end{table}

As shown in Tab. \ref{tab:mapping methods}, the results demonstrate that the representation quality produced by untuned embedding layer projection is significantly inferior. Notably, even after thorough training, the dedicated mapping layer still underperforms compared to our proposed method, demonstrating that the LM Head possesses remarkable generalization capability due to the pre-training process.

\paragraph{\textbf{Cross-modal Demonstration Analysis. }}
Recent studies have shown that in vision-based ICL, multimodal information interaction primarily occurs in the deeper hidden layers of VLMs. In contrast, methods that translate visual content into textual descriptions and use them as input examples allow for earlier cross-modal integration, potentially facilitating more accurate and robust reasoning \cite{zhou2024visual}. Motivated by this observation, we manually annotate four samples from the "difficult" category to construct a set of purely textual examples, consisting of detailed visual descriptions and logical explanations.

However, quantitative analysis reveals a substantial performance gap. Specifically, visual ICL achieves an average nDCG@10 of \textbf{87.8} in reranking tasks, significantly outperforming the text-only variant at \textbf{74.2}. This disparity underscores the fundamental insight that layout-aware reasoning relies on spatial topology, which is inherently compromised by the linearization of text descriptions. In contrast, visual demonstrations preserve these critical 2D patterns (e.g., alignment, grouping), making the visual modality a prerequisite for valid reasoning on complex documents.

\paragraph{\textbf{Selection Strategy Ablation Study.}}
\label{ablationstudy}
To validate the efficacy of our dual-modal selection strategy, we conduct a series of ablation studies comparing four approaches: (1) selection based solely on document similarity, (2) selection based solely on query similarity, (3) a zero-shot baseline without any demonstrations, and (4) our DocRetriever method employing dual-modal similarity. As shown in Tab. \ref{tab:abstudy}, while all demonstration-based approaches improve performance over the zero-shot baseline, single-modal selection strategies consistently underperform the dual-modal approach. This underscores the necessity of integrating both query and document characteristics to achieve optimal reranking fidelity.

\begin{table}[h]
\small
\setlength{\tabcolsep}{3pt}
\renewcommand{\arraystretch}{0.9}
\centering
\begin{tabular}{l|cccc|c}
\toprule
\multirow{2}{*}{\diagbox{Model}{Domain}} 
 & Resear. & Acade. & \multirow{2}{*}{\xspace News\xspace} & \multirow{2}{*}{Others} 
 & \multirow{2}{*}{Average} \\
 & Report & Paper & & & \\
\midrule
Qwen2.5VL & 74.3 & 69.4 & 61.2 & 68.5 & 68.4 \\
Qwen2.5VL(+ Query) & \underline{80.2} & 75.5 & 65.4 & 76.3 & 75.7 \\
Qwen2.5VL(+ Doc.) & 79.1 & \underline{76.1} & \underline{66.3} & \underline{79.8} & \underline{78.1} \\
DocRetriever & \textbf{92.4} & \textbf{90.9} & \textbf{74.8} & \textbf{88.5} & \textbf{87.8} \\
\bottomrule
\end{tabular}
\caption{Ablation study on ICL sampling strategies.}
\label{tab:abstudy}
\end{table}

\paragraph{\textbf{Robustness to Lexical Variations. }}
We further evaluate the impact of lexical variations on different retrieval models. As shown in Tab. \ref{tab:table7}, lexical retrievers are most significantly affected by query rewriting, dropping approximately 8\% in performance. This sensitivity is largely attributed to their heavy reliance on exact term matching, which renders them vulnerable to lexical variations. In contrast, both the dense retrieval model and DocRetriever's hybrid method maintain stable performance, with a decrease of approximately 4\%. This observation aligns with our discussion in Sec. \ref{Hybrid Embedding}. Through a synonym-aware activation mechanism, our sparse embedding bridges the lexical gap by capturing query semantics more robustly, thereby mitigating the negative impact of rephrasing.

\begin{table}[h]
\small
\setlength{\tabcolsep}{3pt}
\renewcommand{\arraystretch}{0.9}
\centering
\begin{tabular}{l|cccc|c}
\toprule
\multirow{2}{*}{\diagbox{Model}{Domain}} 
 & Resear. & Acade. & \multirow{2}{*}{\xspace News\xspace} & \multirow{2}{*}{Others} 
 & \multirow{2}{*}{Average} \\
 & Report & Paper & & & \\
\midrule
TF-IDF & 44.2 & 34.8 & 39.9 & 28.8 & 32.1 \\
ColQ.$_{\mathrm{sparse}}$ & 45.6 & 48.2 & 35.2 & 46.7 & 48.4 \\
Qwen3$_{\mathrm{reranker}}$ & 68.5 & 50.5 & 36.8 & 55.3 & 54.9 \\
ColQ.$_{\mathrm{dense}}$ & \underline{77.6} & \underline{67.2} & \underline{44.3} & \underline{70.8} & \underline{68.5} \\
ColQ.$_{\mathrm{Hybrid}}$ & \textbf{79.4} & \textbf{68.4} & \textbf{50.4} & \textbf{72.5} & \textbf{70.6} \\
\bottomrule
\end{tabular}
\caption{Performance comparison across rephased query.}
\label{tab:table7}
\end{table}

\paragraph{\textbf{Prompt Sensitivity Analysis.}}
\label{sec:prompt_study}
To address concerns on prompt sensitivity of our sparse embedding generation, we evaluate four prompt paradigms on MultiDocR:
\begin{itemize}[leftmargin=*]
    \item \textbf{Compression (Ours):} ``Represent this document in one word:'' 
    \item \textbf{Keyword-centric:} ``What are the keywords of this document?'' 
    \item \textbf{Descriptive:} ``Describe the content of this image:'' 
    \item \textbf{Summarization:} ``Summarize this page:'' 
\end{itemize}

Our Compression strategy achieves the best performance, which achieves nDCG@10 of 0.687. Drawing on the Information Bottleneck Principle~\cite{tishby2000information}, the single-token constraint forces the model to compress visual semantics into highly discriminative vocabulary. Keyword-centric ranks second (0.654), while open-ended prompts (``Describe'': 0.631, ``Summarize'': 0.625) suffer from bias toward high-frequency stopwords, diluting the sparse signal.

\begin{figure}[h]
    \centering
    \includegraphics[page=1, width=0.98\linewidth, clip, trim=6 10 6 6]{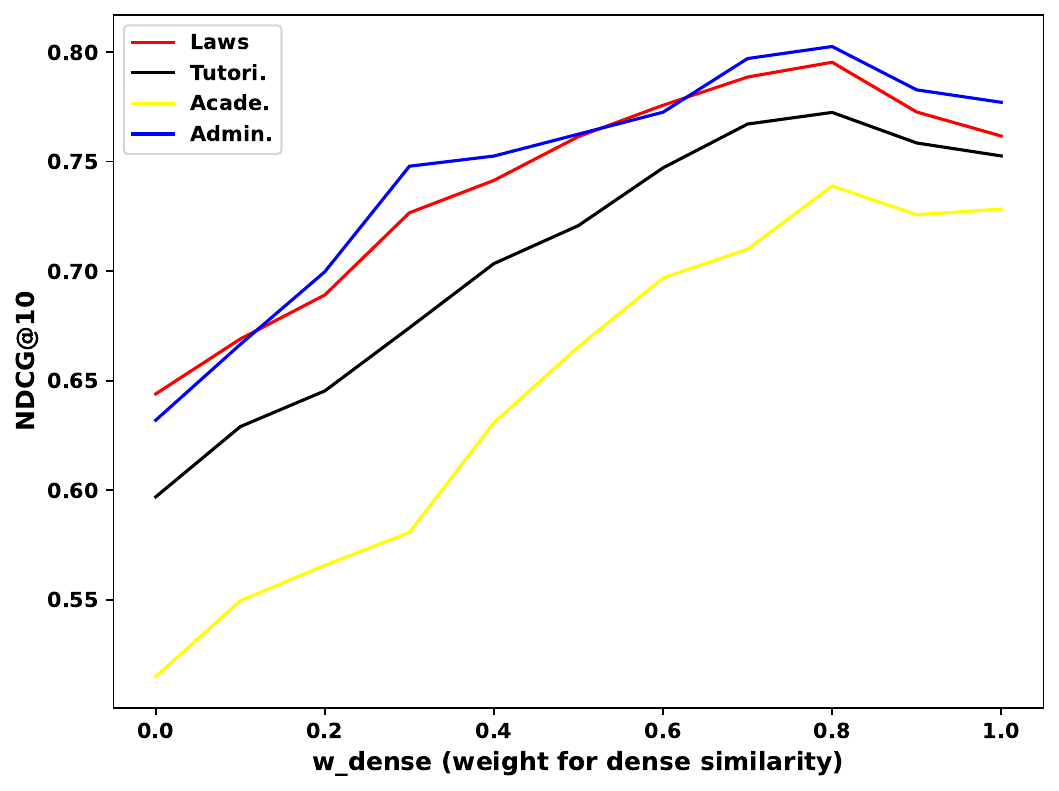}
    \caption{nDCG@10 at different $k$ values vs $w_{dense}$.}
    \label{fig:rationfig}
\end{figure}

\paragraph{\textbf{Parameter Sensitivity of Hybrid Weighting.}}
\label{sec:Parameter Sensitivity}
We examine the sensitivity of the hybrid weighting parameter $\lambda$ to determine the optimal balance between dense and sparse similarity scores. As demonstrated in Fig.~\ref{fig:rationfig}, nDCG@10 peaks at $\lambda = 0.8$, which we adopt as the optimal weighting parameter following established practices~\cite{mandikal2024sparse}. This 4:1 ratio reflects the complementary nature of the two representations: dense embeddings provide broad semantic coverage while sparse embeddings contribute precise lexical matching as a supplementary signal. Notably, performance degrades when $\lambda < 0.6$, confirming that over-reliance on sparse signals alone is insufficient for complex multimodal retrieval.

\section{Conclusion}

In this work, we presented DocRetriever, a plug-and-play framework for multimodal document retrieval with a rigorous evaluation benchmark. Our primary contributions are threefold: First, we proposed a layout-aware hybrid encoding scheme that extracts sparse signals directly from VLM hidden states, significantly boosting the retrieval precision of existing visual encoders. Second, we developed a Reinforced ICL strategy that autonomously synthesizes reasoning-augmented demonstrations, overcoming the scarcity of fine-grained training data and substantially enhancing out-of-distribution generalization. Third, we introduced MultiDocR, a benchmark featuring multi-dimensional taxonomies and 5-level relevance annotations, addressing the limitations of simplistic one-to-one mapping in prior datasets. Future work will explore optimized joint training strategies for hybrid representations and more advanced ICL mechanisms to further improve the accuracy and efficiency of next-generation multimodal retrieval systems.

\begin{acks}
This work was supported in part by the National Natural Science Foundation of China under Grant No.~U25B2064.
\end{acks}

\clearpage

\bibliographystyle{ACM-Reference-Format}
\balance
\bibliography{sample-base}

\appendix

\section{Reinforced ICL Details}
\label{sec:Reinforced ICL Details and Qualitative Analysis}

\begin{table}[h]
\centering
\small
\renewcommand{\arraystretch}{1.2}
\begin{tabular}{ll}
\hline
\textbf{Hyperparameter} & \textbf{Configuration} \\ \hline
Temperature & $0.2$ \\
Top-$p$ & $0.95$ \\
Confidence Threshold & $>0.8$ \\
Max Examples ($k$) & $4$ (2 positive, 2 negative) \\
Verification Strategy & Rotational Tri-party Consensus \\
\hline
\end{tabular}
\caption{Hyperparameters for ICL demonstration generation.}
\label{tab:hyperparams}
\end{table}

\section{Ethical Considerations}
\label{sec:Ethical Considerations}
We address the potential ethical implications of our work, including privacy concerns and algorithmic bias, along with corresponding mitigation strategies.

\subsection{Privacy Concerns}
While our framework enhances models' capability to retrieve relevant information from extensive multimodal documents, this functionality may inadvertently enable the extraction of sensitive personal data (e.g., medical records, financial information). Of particular concern is the potential misuse of this technology for large-scale surveillance systems or unauthorized data mining operations.

\subsection{Fairness and Algorithmic Bias}
When foundation models employed by our framework are trained on datasets lacking adequate representation of demographic characteristics, linguistic variations, and cultural contexts, their outputs may perpetuate societal biases. This could lead to discriminatory decision-making or reinforcement of harmful stereotypes.
\begin{table*}[tb] 
\small
\setlength{\tabcolsep}{2.5pt}
\renewcommand{\arraystretch}{0.75}
    \centering
    \resizebox{0.95\linewidth}{!}{%
  \begin{tabu}{lll|ccccccccc|c}

    \toprule

    \multicolumn{3}{c}{\multirow{2}{*}{\diagbox{Method}{Domain}}} 

    & Econo. &  Bio. & Esg & \multirow{2}{*}{\xspace Arxiv. \xspace} & \multirow{2}{*}{\xspace Chart. \xspace} & \multirow{2}{*}{\xspace Info. \xspace} & \multirow{2}{*}{\xspace MP-Doc. \xspace} & \xspace \multirow{2}{*}{\xspace Plot. \xspace} & \multirow{2}{*}{\xspace Slide. \xspace}   & \multirow{2}{*}{Average} \\
     & & & mics & medical & reports & & & & & & & \xspace \xspace \\
    \midrule

    \parbox[t]{2.5mm}{\multirow{14}{*}{\rotatebox[origin=c]{90}{Recall@3}}} &

    \parbox[t]{1.5mm}{\multirow{5}{*}{\rotatebox[origin=c]{90}{Sparse}}}
    & ColQwen  & 9.5 & 30.3 & 21.5 & 5.6 & 82.5 & 79.5 & 76.6 & 52.6 & 85.4 & 49.3 \\
    && ColPali & 10.5 & 10.4 & 5.6 & 2.1 & 15.9 & 14.2 & 14.9 & 12.2 & 12.8 & 10.9 \\
    && DSE$_{\mathrm{docmatix}}$  & 6.0 & 18.4 & 14.3 & 1.8 & 60.3 & 54.2 & 41.8 & 25.1 & 60.6 & 31.4 \\
    && VisRAG & 7.3 & 17.3 & 12.8 & 3.2 & 61.9 & 42.9 & 36.9 & 28.6 & 75.3 & 31.8 \\
    && VLM2Vec & 6.6 & 15.7 & 11.6 & 2.9 & 56.1 & 38.9 & 33.4 & 25.9 & 68.1 & 28.8 \\
    \tabucline[.2pt on 1pt off 1.5pt]{2-13} \\[-5pt]

    & \parbox[t]{1.5mm}{\multirow{5}{*}{\rotatebox[origin=c]{90}{Dense}}}
    & ColQwen & 15.1 & 45.5 & \underline{42.1} & \underline{8.3} & 84.1 & \underline{96.2} & \underline{95.8} & \underline{72.8} & \underline{96.4} & \underline{61.8} \\
    && ColPali  & \underline{43.4} & 45.5 & 15.1 & 7.5 & 84.1 & 90.4 & 89.5 & 48.7 & 95.8 & 57.8 \\
    && DSE$_{\mathrm{docmatix}}$  & 20.3 & 46.3 & 35.1 & 7.0 & 79.4 & 89.8 & 83.4 & 66.3 & 93.9 & 57.9 \\
    && VisRAG & 15.5 & 40.0 & 34.2 & 6.6 & 77.8 & 92.9 & 82.6 & 69.8 & 90.1 & 56.6 \\
    && VLM2Vec & 16.5 & 39.6 & 28.9 & 5.5 & 78.1 & 89.5 & 78.2 & 69.0 & 89.2 & 54.9 \\
    \tabucline[.2pt on 1pt off 1.5pt]{2-13} \\[-5pt]

    & \parbox[t]{1.5mm}{\multirow{5}{*}{\rotatebox[origin=c]{90}{Ours}}}
    & ColQwen & 17.8 & \textbf{48.9} & \textbf{45.2} & \textbf{9.1} & \textbf{90.5} & \textbf{97.9} & \textbf{95.8} & \textbf{75.9} & \textbf{96.6} & \textbf{64.3} \\
    && ColPali & \textbf{44.7} & 47.0 & 15.1 & 7.9 & \underline{87.3} & 90.3 & 90.2 & 49.1 &95.7 & 58.6 \\
    && DSE$_{\mathrm{docmatix}}$ & 22.6 & \underline{47.2} & 38.8 & 7.6 & 85.0 & 91.9 & 81.8 & 69.0 & 94.2 & 59.8 \\
    && VisRAG & 16.4 & 42.8 & 36.4 & 7.5 & 81.2 & 92.9 & 83.6 & 71.3 & 91.0 & 58.1 \\
    && VLM2Vec & 17.6 & 43.4 & 34.7 & 6.6 & 80.2 & 90.3 & 80.5 & 71.3 & 90.2 & 57.2 \\
    \bottomrule
  \end{tabu}
  }
\caption{Additional results for retrievers on Vidore and VisRAG.}
\label{tab:additional results}
\end{table*}
\subsection{Mitigation Strategies}
To address these challenges, we implement the safeguards:
\begin{itemize}[leftmargin=*]
    \item \textbf{Data Curation}: All benchmark datasets consist exclusively of rigorously vetted, publicly available documents that have undergone thorough anonymization and sensitivity screening.
    \item \textbf{Bias Monitoring}: We establish continuous evaluation protocols to assess fairness metrics and quantify bias dimensions in retrieval outputs.
\end{itemize}
We advocate for the research community to maintain transparency in system capabilities, implement proactive ethical review processes, and foster collaborations for responsible AI development.

\section{Evaluation Metric Definition}
\label{sec:Evaluation Metric Definition}
The Normalized Discounted Cumulative Gain at rank 10 (nDCG@10) is computed as:
\[
\text{DCG}@10 = \sum_{i=1}^{10} \frac{2^{rel_i} - 1}{\log_2(i + 1)}, \quad
\text{nDCG}@10 = \frac{\text{DCG}@10}{\text{IDCG}@10}
\]
where $rel_i$ denotes the relevance score at position $i$, and IDCG@10 is the ideal DCG achievable under perfect ranking.

Our evaluation employs a \textit{weighted} nDCG@10 that prioritizes challenging queries by assigning weights $w_q$ proportional to IDCG@10:
\[
w_q = \frac{\text{IDCG}_q@10}{\sum_{q'}\text{IDCG}_{q'}@10}, \quad
\text{W-nDCG}@10 = \sum_{q=1}^Q w_q \cdot \text{nDCG}_q@10
\]
This design reflects real-world reliability requirements, where queries with more relevant pages are inherently more challenging and thus carry greater weight.

\section{Generalization on External Benchmarks}

\label{sec:Generalization on External Benchmarks}
To validate generalizability, we evaluate on Vidore benchmark-v2~\cite{mace2025vidore} and VisRAG~\cite{yu2024visrag}, restricting to visual dense, sparse, and hybrid encoding as text extraction is unavailable (Tab.~\ref{tab:additional results}).

Baseline models perform poorly on ArxivQA due to its diagram, formula-only content, and strict one-to-one correspondence assumption, so we exclude it from evaluation.

On remaining subsets, hybrid encoding consistently boosts recall across all scenarios, validating generalizability. Performance disparities exist across domains—excelling on VisRAG but struggling on Vidore due to denser terminology and more complex query-document correspondences.

\section{End-to-End RAG Performance}
\label{sec:End-to-End RAG Performance}
To validate the downstream impact of hybrid encoding, we perform an end-to-end evaluation using Qwen2-VL-7B as the reader. Following a standard RAG pipeline, we retrieve the top-4 pages per query and measure Exact Match (EM) and F1 scores, comparing dense-only ColQwen with our hybrid variant ColQwen (hy).

\begin{table}[h]
\centering
\renewcommand{\arraystretch}{1.2}
\begin{tabular}{lcc}
\toprule
\textbf{Retrieval Method} & \textbf{EM (\%)} & \textbf{F1 (\%)} \\
\midrule
ColQwen (Dense-only) & 32.4 & 37.3 \\
\textbf{ColQwen (Hybrid)} & \textbf{36.6} & \textbf{39.4} \\
\bottomrule
\end{tabular}
\caption{End-to-end RAG performance on DocVQA-style tasks (top-4 retrieval).}
\label{tab:end_to_end}
\end{table}

As shown in Tab.~\ref{tab:end_to_end}, hybrid encoding yields a \textbf{4.2\%} absolute gain in EM over the dense-only baseline, confirming that superior page-level retrieval directly improves downstream answer accuracy.

\section{Comparison with PromptReps}
\label{sec:Methodological Comparisons Clarifications}
DocRetriever and PromptReps~\cite{zhuang2024promptreps} both use LLMs to generate dense and sparse embeddings in a single forward pass, but differ across three dimensions:

\paragraph{\textbf{Encoding Mechanism.}} DocRetriever employs architecture-aware strategies per backbone (e.g., chunk-level processing with max-pooling for ColQwen), whereas PromptReps uniformly uses last hidden states without architectural differentiation.

\paragraph{\textbf{System Adaptability.}} DocRetriever is a plug-and-play enhancement for existing VLM retrievers, ensuring robustness across backbones. PromptReps targets LLM base models with prompt-sensitive dense embeddings~\cite{tao2024llms}, requiring case-specific engineering that hinders a universal framework.

\paragraph{\textbf{Experimental Findings.}}
\begin{itemize}[leftmargin=*]
\item \textbf{Weight Fusion:} DocRetriever identifies an optimal 4:1 dense-sparse ratio for VLMs; PromptReps finds 1:1 optimal for LLMs, reflecting differential retrieval-oriented fine-tuning.
\item \textbf{Sparse Efficacy:} DocRetriever's sparse embeddings surpass BM25 independently, whereas PromptReps' sparse representations generally underperform BM25 and rely on dense-sparse synergy.
\end{itemize}

\end{document}